\documentclass[sigconf]{acmart}

\setcopyright{acmlicensed}
\copyrightyear{2018}
\acmYear{2018}
\acmDOI{XXXXXXX.XXXXXXX}
\acmConference[Conference acronym 'XX]{Make sure to enter the correct
  conference title from your rights confirmation email}{June 03--05,
  2018}{Woodstock, NY}

\usepackage{multirow}
\usepackage[utf8]{inputenc}
\usepackage{xcolor}
\usepackage{colortbl}
\usepackage{enumitem}

\usepackage{booktabs}
\usepackage{graphicx}
\usepackage[table]{xcolor}

\definecolor{gain1}{RGB}{245,249,246}
\definecolor{gain2}{RGB}{231,242,234}
\definecolor{gain3}{RGB}{211,233,217}
\definecolor{gain4}{RGB}{181,216,191}
\definecolor{gain5}{RGB}{151,199,164}
\definecolor{lightgray}{RGB}{242,242,242}

\definecolor{loss1}{RGB}{253,244,244}
\definecolor{loss2}{RGB}{249,228,228}
\definecolor{loss3}{RGB}{243,208,208}

\newcommand{\gaincell}[2]{\cellcolor{gain#1}#2}

\usepackage{float}
\begin{document}

\title{HiGenTrans: Hierarchical LLM Knowledge Transfer for Generative Recommendation}

\author{Zhuohang Jiang}
\authornote{Work done during an internship at Kuaishou Technology.}
\authornote{Authors contributed equally to this research.}
\affiliation{%
  \institution{The Hong Kong Polytechnic University}
  \city{Hong Kong SAR}
  \country{China}
}
\email{zhuohang.jiang@connect.polyu.hk}

\author{Yuxin Chen}
\authornotemark[2]
\authornote{Corresponding Author.}
\affiliation{%
  \institution{Kuaishou Technology}
  \city{Beijing}
  \country{China}
}
\email{chenyuxin06@kuaishou.com}

\author{Haohao Qu}
\affiliation{%
  \institution{The Hong Kong Polytechnic University}
  \city{Hong Kong SAR}
  \country{China}
}
\email{haohao.qu@connect.polyu.hk}

\author{Zefei Luo}
\author{Yunlong Wang}
\author{Xu Yuan}
\affiliation{%
  \institution{Kuaishou Technology}
  \city{Beijing}
  \country{China}
}

\author{Haojie Ding}
\author{Boheng Zhang}
\author{Fan Yan}
\affiliation{%
  \institution{Kuaishou Technology}
  \city{Beijing}
  \country{China}
}

\author{Hongyang Wang}
\author{Jun Wang}
\affiliation{%
  \institution{Kuaishou Technology}
  \city{Beijing}
  \country{China}
}

\author{Zhongbo Sun}
\affiliation{%
  \institution{Kuaishou Technology}
  \city{Beijing}
  \country{China}
}
\email{szb17@tsinghua.org.cn}

\author{Wenqi Fan}
\authornotemark[3]
\affiliation{%
  \institution{The Hong Kong Polytechnic University}
  \city{Hong Kong SAR}
  \country{China}
}
\email{wenqifan03@gmail.com}

\author{Qing Li}
\authornotemark[3]
\affiliation{%
  \institution{The Hong Kong Polytechnic University}
  \city{Hong Kong SAR}
  \country{China}
}
\email{csqli@comp.polyu.edu.hk}

\author{Wenwu Ou}
\affiliation{%
  \institution{Kuaishou Technology}
  \city{Beijing}
  \country{China}
}
\email{ouwenwu@gmail.com}

\author{Peng Jiang}
\affiliation{%
  \institution{Kuaishou Technology}
  \city{Beijing}
  \country{China}
}
\email{jiangpeng11@kuaishou.com}

\renewcommand{\shortauthors}{Zhuohang Jiang et al.}

\begin{abstract}

Generative recommendation promises to simplify conventional multi-stage recommender systems by generating item Semantic IDs (SIDs) autoregressively.
However, two obstacles limit its effectiveness in industrial settings: interaction-only supervision underuses the semantic knowledge encoded in pretrained language models, while beam-search likelihood ranks token sequences rather than directly estimating item relevance.
To address these limitations, we propose \ourname{}, a three-level framework that introduces LLM knowledge at the input,
representation, and decision levels.
First, \textbf{\MODa{} (GSP)} adds an LLM-derived user representation to the beginning-of-sequence embedding, providing a semantic prior before SID generation.
Second, \textbf{\MODb{} (HRT)} aligns intermediate LLM representations with depth-matched GenRec layers through mean-centered projections and gated residual connections.
Finally, \textbf{\MODc{} (MDC)} uses a lightweight mixture-of-experts calibrator to combine semantic relevance, generation confidence, and candidate-set context when reranking beam-search outputs.
All semantic features are computed only from the observable interaction prefix, preventing target leakage, and can be cached offline.
Experiments on three public benchmarks show that \ourname{} improves Recall@10 by up to \textbf{12.9\%} and NDCG@10 by up to \textbf{33.7\%} over the strongest baselines.
These results demonstrate the value of transferring LLM knowledge at complementary stages of the generative recommendation pipeline.
\end{abstract}




\newcommand{\ourname}{HiGenTrans}
\newcommand{\MODa}{Global Semantic Priming}
\newcommand{\MODb}{Hierarchical Representation Transfer}
\newcommand{\MODc}{Multi-Signal Decision Calibration}

\maketitle

\section{Introduction}

Recommender systems are central to \textbf{Web} applications such as
e-commerce, job search, and social media, where they help users navigate rapidly growing item spaces~\cite{kang2018self,fan2019graph,fan2020graph}.
Conventional industrial systems commonly organize retrieval, ranking, and reranking as separately optimized stages
Conventional industrial systems commonly organize multi-stage discriminative architectures~\cite{cheng2026choruscvr,qu2024ssd4rec}, where candidate items are progressively filtered through retrieval~\cite{zhang2026onemall}, ranking~\cite{xia2026harmonrank}, and reranking~\cite{jiang2026b} stages.
Although this paradigm has achieved widespread success in large-scale industrial recommender systems~\cite{pi2020search}, different stages often rely on relatively independent modeling objectives and optimization strategies.
Such a multi-stage architecture increases system design, training, and maintenance complexity, while making it challenging to jointly optimize user preference modeling and final recommendation decisions.
In recent years, generative recommendation (GenRec)~\cite{deng2025onerec,zhang2026onemall,qu2026diffusion} has emerged as a promising alternative paradigm.
Unlike traditional discriminative approaches, GenRec reformulates recommendation as an autoregressive sequence generation task, typically encoding items into discrete Semantic IDs (SIDs)~\cite{hu2026stop,pan2026beyond,qu2025tokenrec} and employing Transformer-based models to generate target items token by token based on users' historical interaction sequences.
Representative studies, such as TIGER~\cite{rajput2023recommender} and OneRec~\cite{deng2025onerec}, have demonstrated the potential of generative modeling for sequential recommendation, opening up new directions for building more unified and scalable recommendation models.

\begin{figure}[t]
    \centering
    \includegraphics[width=\linewidth]{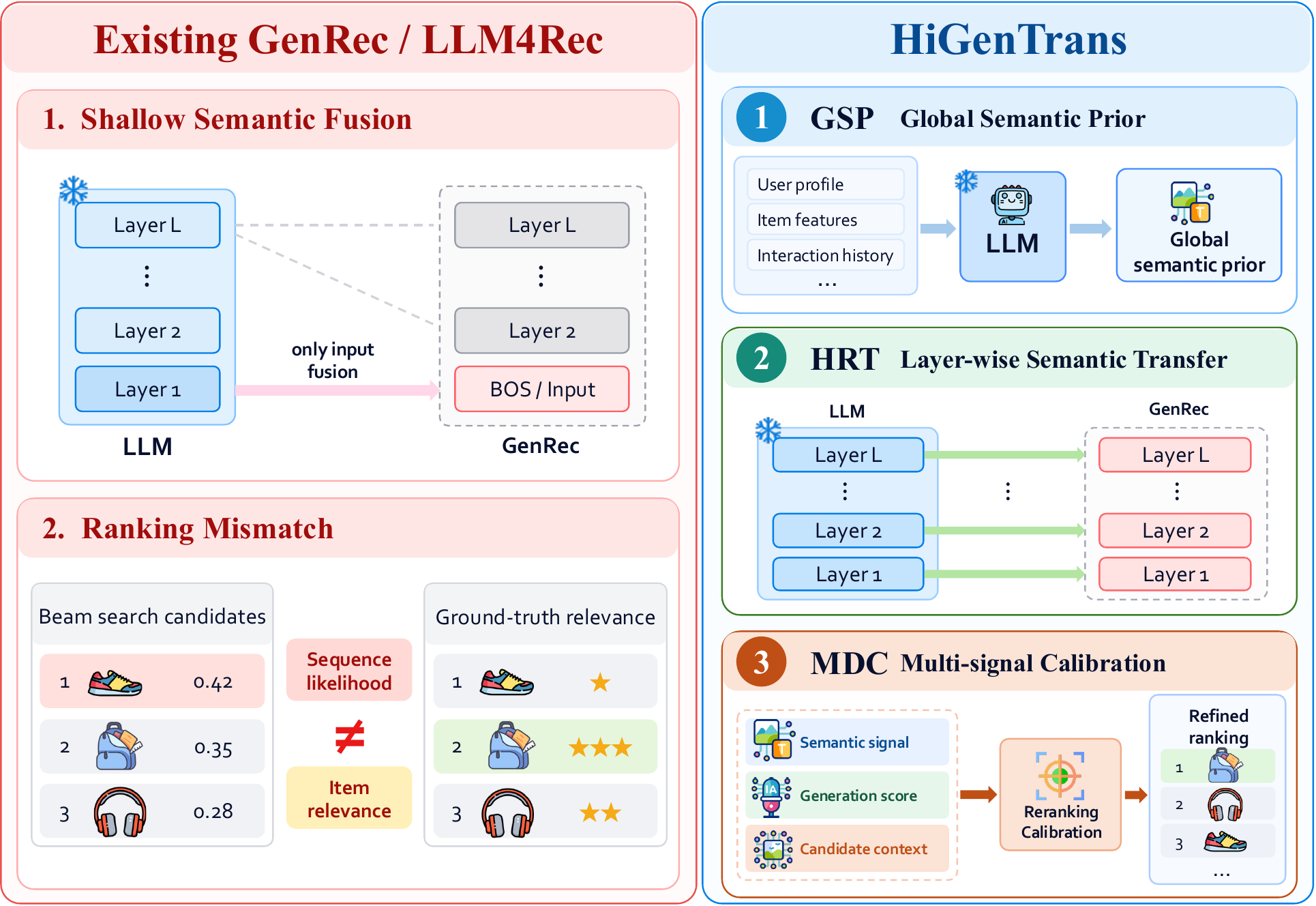}
    \caption{
     Motivation and overview of HiGenTrans. Existing GenRec and LLM4Rec methods suffer from shallow semantic fusion and misalignment between generation likelihood and item relevance. HiGenTrans addresses these limitations through Global Semantic Prior (GSP), Hierarchical Representation Transfer (HRT), and Multi-signal Decision Calibration (MDC), enabling richer semantic integration and more accurate recommendation ranking.
    }
    \label{fig:motivation}
    \vskip -0.1in
\end{figure}

Despite the progress of GenRec in advancing recommendation paradigms, as illustrated in Fig.~\ref{fig:motivation}, they also encounter several limitations.
The first limitation arises from how GenRec uses semantic knowledge~\cite{zhai2024actions}.
Interaction supervision is effective for learning collaborative regularities, but it can be sparse and does not directly expose the textual relations among items or the high-level intent underlying a history. Pretrained LLMs provide complementary semantic and contextual knowledge~\cite{jiang2025qa,jiang2026superglasses,dai2025large,wang2025knowledge,yuan2026ai,ning2025survey}; accordingly, recent LLM-enhanced recommenders derive interest descriptions or semantic features from item metadata~\cite{hu2026retrieval,zhang2025towards,lee2025gram,hong2025eager}. Most fusion strategies, however, compress this information into one or a few input embeddings, such as prompts, prefixes, or beginning-of-sequence (BOS) tokens~\cite{chen2026onesearch}.
Because representations at different Transformer depths encode complementary abstractions~\cite{jawahar2019does,tenney2019bert,chuang2024dola,jiang2026atomic}, a single input-level injection leaves much of the LLM hierarchy unused and requires the recommendation backbone to preserve one semantic vector through all subsequent layers.
The second limitation concerns the decision rule.
Beam search orders SID sequences by autoregressive likelihood, which is affected by tokenization, prefix probabilities, and decoding dynamics. This sequence-level likelihood need not equal the relevance of the decoded item to the user~\cite{wang2026onelive}.
A relevant item can therefore appear in the candidate set but occupy a poor rank.
Correcting this mismatch requires candidate-level evidence beyond the raw generation score, including semantic relevance, token-level confidence, and each candidate's position relative to the rest of the beam.
These two limitations suggest that LLM knowledge should enter GenRec not only once at the input, but also within its representation hierarchy and at its final decision stage.

We therefore propose \textbf{HiGenTrans}, a \textbf{Hi}erarchical LLM-to-\textbf{Gen}Rec Knowledge-\textbf{Trans}fer framework, which integrates three interactive modules, namely \emph{\MODa{} (GSP)} for prior kownledge encoding, \emph{\MODb{} (HRT)} for layer-wise representation adaptation, and \emph{\MODc{} (MDC)} for efficient feature mixing.
Specifically, the GSP module projects a frozen LLM representation of the observable user-history prefix into the BOS embedding, giving the generator a global semantic prior.
Moreover, HRT then mean-centers, projects, and injects intermediate LLM representations into depth-matched GenRec layers through learned residual gates.
GSP and HRT preserve the collaborative backbone while exposing it to semantic information both before and during SID generation.
Finally, MDC applies a lightweight mixture-of-experts calibrator to semantic, generative, and candidate-context features and reranks only the beam-generated items.
Thus, GSP initializes \emph{what the model should attend to}, HRT enriches \emph{how the history is represented}, and MDC calibrates \emph{which generated candidates should be recommended}. All LLM representations are computed from the prefix available at the corresponding prediction step, which prevents target leakage, and the frozen encoder permits offline feature caching.

Considering the stringent storage and inference efficiency requirements of industrial recommender systems, we develop an efficient deployment strategy for large-scale deployment.
Since HRT relies on semantic representations extracted from multiple intermediate LLM layers, directly storing and transmitting these high-dimensional features would introduce substantial system overhead.
To address this issue, we introduce a vector quantization-based feature compression mechanism that quantizes the intermediate LLM representations at different layers into compact Semantic IDs and reconstructs the corresponding features through a shared codebook.
By performing LLM feature extraction and quantization encoding entirely offline, the online inference stage only requires lightweight feature reconstruction without invoking the LLM.

We conduct extensive experiments on three public recommendation datasets, namely Beauty, Sports, and Toys, to comprehensively evaluate the effectiveness of HiGenTrans. Experimental results demonstrate that HiGenTrans consistently outperforms existing baseline methods across all evaluation metrics on all three datasets, achieving substantial performance improvements. Specifically, HiGenTrans improves Recall@10 by 2.69\%, 5.69\%, and 12.94\% on the Beauty, Sports, and Toys datasets, respectively.
The ablations show complementary gains from semantic transfer (GSP and HRT together) and decision calibration (MDC), while experiments in two industrial domains support the applicability of hierarchical semantic fusion in production settings.

Our main contributions are summarized as follows:

\begin{itemize}[leftmargin=*]
    \item We propose \textbf{\ourname{}}, a three-level framework that transfers LLM knowledge at the input, internal representation, and decision stages of generative recommendation.

    \item We introduce \textbf{\MODa{} (GSP)} and \textbf{\MODb{} (HRT)}, which combine prefix-safe global initialization with mean-centered, gated fusion of depth-matched LLM representations.

    \item We develop \textbf{\MODc{} (MDC)}, a lightweight candidate-level calibrator that uses a mixture of experts to integrate semantic, generative, and contextual evidence.

    \item Experiments on three public benchmarks and two industrial domains demonstrate consistent improvements, while ablation and segment-level analyses characterize the contributions of semantic transfer and decision calibration.

\end{itemize}

\section{Related Works}

\subsection{Generative Recommendation}

Generative recommendation (GenRec) casts item retrieval as conditional sequence generation, replacing a catalog-wide scoring operation with autoregressive decoding over discrete item representations~\cite{rajput2023recommender}.
TIGER~\cite{rajput2023recommender} pioneered this paradigm by representing items with hierarchical Semantic IDs (SIDs), allowing recommendation to be performed through discrete token generation.
Subsequent work has mainly advanced GenRec along three directions.
First, tokenization methods improve the correspondence between discrete codes and recommendation structure: LETTER~\cite{wang2024learnable} jointly models semantic, collaborative, and code-diversity objectives, TokenRec~\cite{qu2025tokenrec} learns ID tokenization for LLM-based generation, and qualification-aware or adaptive collision handling makes SIDs more suitable for large catalogs~\cite{hu2026stop,pan2026beyond}.
Second, model architectures strengthen the interaction between behavioral and semantic signals~\cite{xiao2025progressive}.
EAGER~\cite{wang2024eager} uses two streams for behavior--semantic collaboration,
RPG~\cite{hou2025generating} generates long SIDs in parallel, and GRLM~\cite{zhang2026unleashing} uses structured term identifiers to expose the recommendation capability of LLM-based generators.
Third, industrial systems seek unified or scalable generation, as illustrated by OneRec~\cite{deng2025onerec}, OneMall~\cite{zhang2026onemall}, and OneLive~\cite{wang2026onelive}.
These studies improve \emph{what} is generated or \emph{how} a generator is trained, but they do not fully address where external semantic knowledge should enter the recommendation pipeline.
Most semantic information is encoded into item tokens, fused at the input, or distilled into the generator.
In contrast, \ourname{} keeps the SID-generation backbone while transferring prefix-conditioned LLM knowledge at three distinct locations: GSP supplies a global prior before decoding, HRT injects depth-specific features inside the backbone, and MDC calibrates the ordering of the resulting candidate set.
This design separates semantic augmentation of candidate generation from candidate-level decision calibration.

\subsection{LLM-Enhanced Recommendation}

LLMs have been incorporated into recommender systems as semantic encoders, reasoning modules, or recommendation backbones~\cite{kim2024large,qu2026diffusion,zhao2024recommender}.
One line of work adapts an LLM to collaborative signals. CLLM4Rec~\cite{zhu2024collaborative} extends the LLM vocabulary with user and item tokens, while CoLLM~\cite{zhang2025collm}  connects collaborative representations or learned identifiers to language models. Another line transfers language-model knowledge to comparatively lightweight recommenders.
LLM-ESR~\cite{liu2024llm} enhances ID-based sequential recommendation with LLM-derived semantics, IDLE-Adapter~\cite{yu2024break} aligns representations and output distributions, and distribution-matching approaches explicitly bridge collaborative and language spaces~\cite{zhang2025towards}. IMFuse~\cite{zheng2026imfuse} further shows that intermediate LLM layers contain complementary information that can be aggregated for sequential recommendation.
Our setting differs in both transfer direction and granularity. Rather than turning the LLM itself into the online recommender, \ourname{} uses a frozen LLM as an offline semantic feature provider for a generative recommendation backbone. Moreover, it does not collapse all intermediate layers into one representation. Motivated by evidence that Transformer layers encode different levels of abstraction~\cite{jawahar2019does,tenney2019bert,chuang2024dola}, HRT preserves the ordering of selected LLM layers and maps them monotonically to GenRec depth. Mean-centered projections address the mismatch between the two representation spaces, and learned residual gates let recommendation supervision determine the strength of each transfer. The resulting model therefore complements input-only, final-layer, and layer-aggregation approaches with explicit depth-to-depth transfer.
\begin{figure*}[t]
    \centering
    \includegraphics[width=\linewidth]{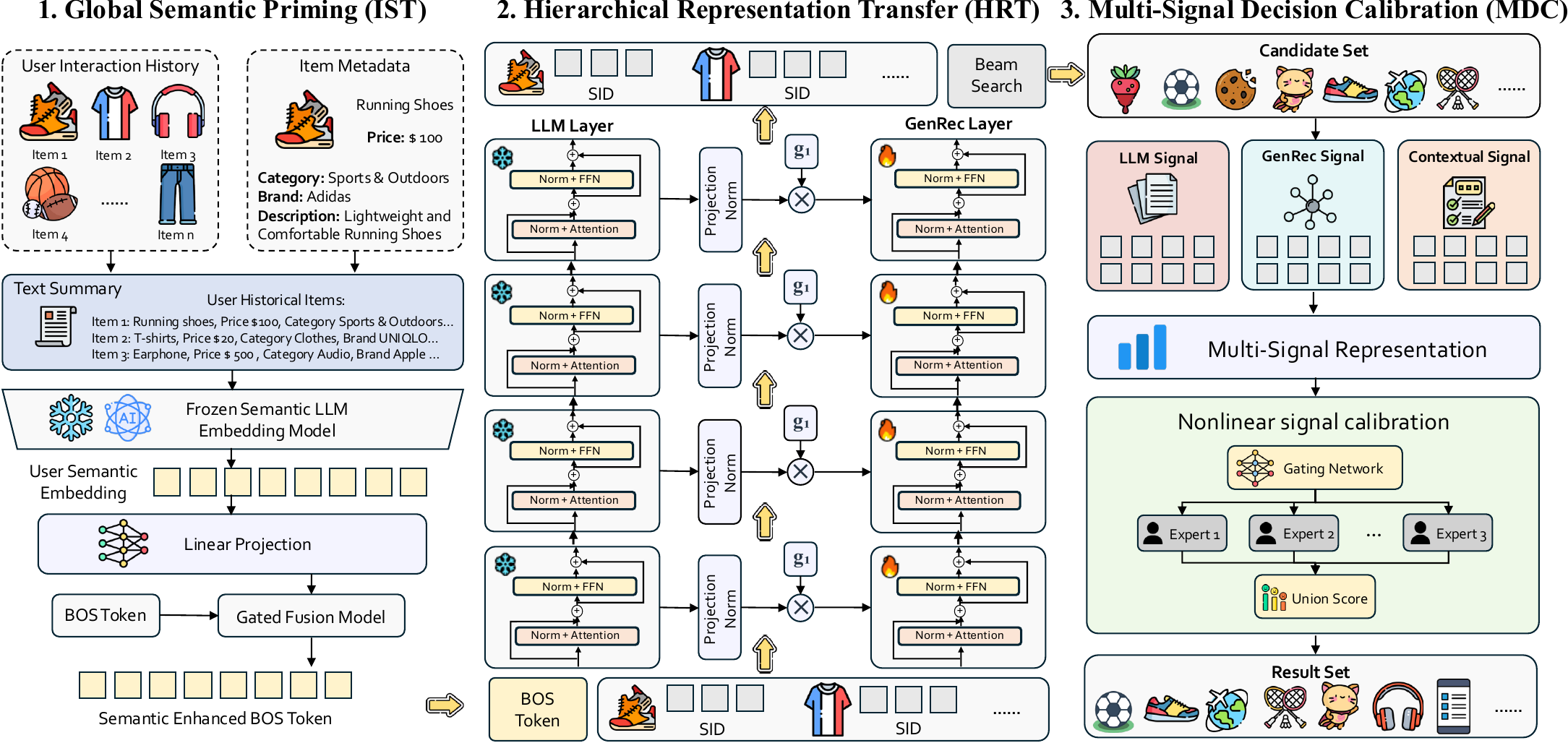}
    \caption{
    Overview of \ourname{}, a three-level semantic knowledge transfer framework for generative recommendation. \MODa{} (GSP) incorporates global user semantics into the BOS token. \MODb{} (HRT) injects multi-level LLM representations into the generative backbone. \MODc{} (MDC) employs MoE-based multi-signal calibration to refine candidate rankings.
    }
    \label{fig:methodology}
\end{figure*}

\section{Methodology}

\subsection{Framework Overview}

Existing generative recommendation models predict item sequences autoregressively using Semantic IDs, but often underutilize the hierarchical semantic knowledge of pretrained LLMs and rely on generative likelihoods that may not align with item-level ranking relevance. To address these limitations, we propose \ourname{}, a framework that transfers LLM semantic knowledge into generative recommendation at multiple levels, as illustrated in Fig.~\ref{fig:methodology}.

\noindent\textbf{Notation.} Let $\mathcal{H}_{u,\leq t}=(i_1,\ldots,i_t)$ denote the observable interaction history of user $u$ up to step $t$, and let $i_{t+1}$ be the target item. Each item $i$ is represented by a sequence of $K$ Semantic ID (SID) tokens $\mathbf{y}_i=(y_{i,1},\ldots,y_{i,K})$, where $\mathbf{y}_{i,<k}=(y_{i,1},\ldots,y_{i,k-1})$. We use $\mathcal{D}_{\mathrm{gen}}=\{(u,t,i_{t+1})\}$ to denote the collection of next-item training instances. A generative recommender factorizes the next-item probability as
\begin{equation}
p_{\theta}(\mathbf{y}_{i_{t+1}}\mid\mathcal{H}_{u,\leq t})
=
\prod_{k=1}^{K}
p_{\theta}
\left(
y_{i_{t+1},k}
\mid
\mathcal{H}_{u,\leq t},
\mathbf{y}_{i_{t+1},<k}
\right).
\label{eq:gen_factorization}
\end{equation}

Our framework enhances this generative process through three complementary components. \textbf{\MODa{} (GSP)} incorporates global user semantics to guide the generation process. \textbf{\MODb{} (HRT)} injects intermediate LLM representations into depth-matched layers of the generative backbone, enabling hierarchical semantic knowledge transfer. \textbf{\MODc{} (MDC)} integrates semantic, generative, and contextual signals to refine candidate rankings. The pretrained LLM remains frozen and is used only for prefix-safe feature precomputation, while the generative backbone and fusion parameters are jointly optimized. Finally, MDC is trained on candidates produced by the trained generator to improve item-level ranking.
\subsection{\MODa{}}

GenRec processes discrete interaction tokens, which preserve collaborative structure but may omit the high-level intent expressed by item titles and categories.
The \MODa{} (GSP) module supplies this information before autoregressive SID generation begins.
For each training instance $(u,t,i_{t+1})$, we serialize only the observable prefix $\mathcal{H}_{u,\leq t}$ with available item metadata. A frozen semantic encoder produces a global prefix representation

\begin{equation}
\mathbf{s}_{u,t}
=
E_{\mathrm{LLM}}
\left(
\mathrm{Text}(\mathcal{H}_{u,\leq t})
\right)
\in\mathbb{R}^{d_{\mathrm{LLM}}},
\label{eq:input_semantic}
\end{equation}
where $E_{\mathrm{LLM}}(\cdot)$ denotes the frozen encoder together with its fixed sequence-pooling operation, and $d_{\mathrm{LLM}}$ is its output dimension. We project this vector into the $d_{\mathrm{rec}}$-dimensional GenRec space and add it to the beginning-of-sequence representation:

\begin{equation}
\widetilde{\mathbf{h}}_{u,t}^{\mathrm{BOS}}
=
\mathbf{h}^{\mathrm{BOS}}
+
W_{\mathrm{GSP}}\mathbf{s}_{u,t},
\quad
W_{\mathrm{GSP}}\in
\mathbb{R}^{d_{\mathrm{rec}}\times d_{\mathrm{LLM}}}.
\label{eq:input_transfer}
\end{equation}

Thus, the BOS token becomes a semantic prior conditioned on the same history available to the recommender. Computing $\mathbf{s}_{u,t}$ from $\mathcal{H}_{u,\leq t}$ rather than the complete sequence is essential: using $i_{t+1}$ or later interactions would leak the target into the input. Because the semantic encoder is frozen, all prefix representations can be computed and cached before GenRec training.

\subsection{\MODb{}}

GSP provides the generator with a global semantic prior,but a single input-level representation cannot fully preserve the complementary abstractions encoded at different LLM depths~\cite{jawahar2019does,tenney2019bert,chuang2024dola}.
Moreover, its influence may weaken as it propagates through the generative backbone.
We therefore propose \textbf{\MODb{} (HRT)}, which transfers depth-specific semantic knowledge directly into the backbone's internal representations. HRT contains three components: (i) \emph{multi-level semantic extraction}, which obtains prefix-conditioned features from representative LLM layers; (ii) \emph{representation adaptation}, which mean-centers, projects, and normalizes each feature into the GenRec space; and (iii) \emph{gated layer-wise fusion}, which injects the adapted features into depth-matched backbone layers through residual connections. The resulting transfer parameters and the GenRec backbone are then trained jointly using the autoregressive recommendation objective. The following subsections describe these components and the joint optimization procedure in detail.

\subsubsection{\textbf{Multi-level semantic extraction}}
Let $\mathcal{S}=\{l_1,\ldots,l_M\}$ be the selected LLM layers.
Generally, pretrained LLMs encode semantic information at different levels of abstraction across their Transformer layers~\cite{jawahar2019does,tenney2019bert}.
To exploit this hierarchical knowledge, we extract hidden representations from a set of representative layers of the frozen semantic encoder.
Specifically, for each observable prefix, we extract

\begin{equation}
\mathbf{e}_{u,t}^{(l)}
=
E_{\mathrm{LLM}}^{(l)}
\left(
\mathrm{Text}(\mathcal{H}_{u,\leq t})
\right)
\in\mathbb{R}^{d_{\mathrm{LLM}}},
\quad l\in\mathcal{S},
\label{eq:semantic_encoding}
\end{equation}
where $E_{\mathrm{LLM}}^{(l)}(\cdot)$ applies the same fixed pooling operation to the token representations returned by LLM layer $l$.
In our implementation, we use \emph{Qwen3-Embedding-8B} as the frozen encoder and choose four layers distributed across its depth.
An order-preserving injective mapping $\pi:\mathcal{S}\rightarrow\{1,\ldots,L_{\mathrm{rec}}\}$ associates each selected LLM layer with a GenRec layer of similar relative depth, where $L_{\mathrm{rec}}$ is the number of backbone layers. Thus, $l_a<l_b$ implies $\pi(l_a)<\pi(l_b)$, preventing multiple semantic features from being assigned ambiguously to the same backbone layer.

\subsubsection{\textbf{Representation adaptation}}
The LLM and GenRec feature spaces differ in both dimension and distribution~\cite{zhang2025towards}. Moreover, LLM embeddings can be anisotropic, so a direction shared by most histories may dominate direct residual injection.

To mitigate these issues, we introduce a lightweight representation adaptation mechanism that combines mean centering, linear projection, and normalization.
For each selected layer, we estimate the training-set mean using only training prefixes:

\begin{equation}
\boldsymbol{\mu}^{(l)}
=
\frac{1}{|\mathcal{D}_{\mathrm{gen}}|}
\sum_{(u,t,i_{t+1})\in\mathcal{D}_{\mathrm{gen}}}
\mathbf{e}_{u,t}^{(l)}.
\label{eq:semantic_mean}
\end{equation}

The centered feature is projected and normalized as

\begin{equation}
\mathbf{z}_{u,t}^{(l)}
=
\mathrm{RMSNorm}
\left[
W_l
\left(
\mathbf{e}_{u,t}^{(l)}-\boldsymbol{\mu}^{(l)}
\right)
\right],
\quad
W_l\in\mathbb{R}^{d_{\mathrm{rec}}\times d_{\mathrm{LLM}}}.
\label{eq:semantic_projection}
\end{equation}
where $\boldsymbol{\mu}^{(l)}$ denotes the mean semantic representation estimated from the training data at layer $l$.
Mean subtraction is performed in FP32 to avoid cancellation errors under mixed-precision training. Centering removes shared components, the learned matrix aligns the two feature spaces, and RMS normalization controls the scale of the injected vector.

\subsubsection{\textbf{Gated layer-wise fusion}}
Let $\mathbf{h}_{u,t,k}^{(\pi(l))}\in\mathbb{R}^{d_{\mathrm{rec}}}$ be the hidden state at SID decoding step $k$ immediately before GenRec layer $\pi(l)$. HRT applies the same prefix-conditioned semantic feature at every decoding step through

\begin{equation}
\widetilde{\mathbf{h}}_{u,t,k}^{(\pi(l))}
=
\mathbf{h}_{u,t,k}^{(\pi(l))}
+
g_l\mathbf{z}_{u,t}^{(l)},
\quad
g_l\in\mathbb{R},
\ l\in\mathcal{S},
\ k\in\{1,\ldots,K\},
\label{eq:hier_injection}
\end{equation}

where $g_l$ is a trainable layer-specific scalar gate. The residual path preserves the backbone's collaborative representation, while depth-matched injection lets earlier semantic features undergo further transformation and supplies later features closer to the output.

\subsubsection{\textbf{Joint generative optimization}}
Let
$\mathcal{E}_{u,t}=\{\mathbf{s}_{u,t}\}\cup\{\mathbf{z}_{u,t}^{(l)}:l\in\mathcal{S}\}$
collect the global and layer-wise semantic features used by GSP and HRT. The GenRec backbone and the transfer parameters are jointly optimized with

\begin{equation}
\mathcal{L}_{\mathrm{gen}}
=
-\sum_{(u,t,i_{t+1})\in\mathcal{D}_{\mathrm{gen}}}
\sum_{k=1}^{K}
\log p_{\theta}
\left(
y_{i_{t+1},k}
\mid
\mathcal{H}_{u,\leq t},
\mathbf{y}_{i_{t+1},<k},
\mathcal{E}_{u,t}
\right).
\label{eq:gen_loss}
\end{equation}

Here, $\theta$ collects the trainable GenRec, projection, and fusion parameters; the LLM remains frozen. We use no auxiliary representation-alignment loss, so recommendation supervision determines how the transferred features are adapted.

\subsection{\MODc{}}

Although GSP and HRT improve candidate generation, beam-search scores measure the conditional likelihood of SID sequences rather than item-level recommendation relevance~\cite{rajput2023recommender,deng2025onerec}. Consequently, candidates with similar generation scores may differ substantially in semantic relevance and relative quality within the beam. We therefore propose \textbf{\MODc{} (MDC)}, which calibrates the generator's outputs at the decision level. MDC contains three components: (i) \emph{multi-signal candidate representation}, which combines LLM semantic evidence, GenRec confidence statistics, and candidate-set context; (ii) \emph{nonlinear signal calibration}, which uses an input-dependent mixture of experts to estimate candidate relevance; and (iii) \emph{calibrated candidate ranking}, which reorders and truncates the beam according to the calibrated scores. The following subsections present these components and the corresponding training objective.

\subsubsection{\textbf{Multi-signal candidate representation}}
For user prefix $\mathcal{H}_{u,\leq t}$, beam search generates SID sequences that are mapped to an item-level candidate set
$\mathcal{C}_{u,t}=\{c_1,\ldots,c_B\}$,
where $B$ is the beam size and $c_i$ denotes a decoded item candidate. For every $c_i\in\mathcal{C}_{u,t}$, we construct

\begin{equation}
\mathbf{x}_{u,t,i}
=
\phi
\left(
\mathcal{H}_{u,\leq t},
c_i,
\mathcal{C}_{u,t},
\mathcal{E}_{u,t}
\right)
\in\mathbb{R}^{d_f}.
\label{eq:rank_features}
\end{equation}

The feature extractor $\phi(\cdot)$ combines three signal families: (i) \emph{LLM semantic signals}, which measure consistency between a candidate and the user's observed history; (ii) \emph{GenRec signals}, including sequence likelihood, beam rank, and token-level confidence statistics; and (iii) \emph{candidate-context signals}, which describe a candidate relative to the score distribution of $\mathcal{C}_{u,t}$. We retain raw features and append candidate-set-wise standardized versions, thereby preserving both absolute confidence and within-beam differences.

\subsubsection{\textbf{Nonlinear Signal Calibration}}

To capture nonlinear interactions among heterogeneous signals, MDC employs $J$ MLP experts with an input-dependent gating network. Given the candidate feature vector $\mathbf{x}_{u,t,i}$, the calibrated relevance score is computed as
\begin{equation}
s_{u,t,i}
=
\sigma\left(
\sum_{j=1}^{J}
\left[\operatorname{Softmax}
(W_g\mathbf{x}_{u,t,i}+\mathbf{b}_g)\right]_j
f_j(\mathbf{x}_{u,t,i})
\right),
\label{eq:ranking_score}
\end{equation}
where $f_j$ denotes the $j$-th expert, and the Softmax operation produces input-dependent gating weights for expert aggregation.

With the generator frozen, MDC is optimized over generated candidates using binary cross-entropy:
\begin{equation}
\mathcal{L}_{\mathrm{MDC}}
=
\frac{1}{|\mathcal{D}_{\mathrm{cand}}|}
\sum_{(u,t,i)\in\mathcal{D}_{\mathrm{cand}}}
\ell_{\mathrm{BCE}}(r_{u,t,i},s_{u,t,i}),
\label{eq:rank_loss2}
\end{equation}
where $r_{u,t,i}=\mathbb{I}[i=i_{t+1}]$ indicates whether a candidate matches the ground-truth item, and $\mathcal{D}_{\mathrm{cand}}$ denotes the candidate training instances. This two-stage optimization enables refined candidate ranking without modifying the underlying generator.

\subsubsection{\textbf{Calibrated candidate ranking}}
At inference time, the recommendation list contains the $N\leq B$ candidates with the highest calibrated scores:

\begin{equation}
\mathcal{R}_{u,t}
=
\mathrm{TopK}_{N}
\left(
\left\{
(c_i,s_{u,t,i})
\mid
c_i\in\mathcal{C}_{u,t}
\right\}
\right).
\label{eq:final_ranking}
\end{equation}

MDC can change only the ordering and truncation of $\mathcal{C}_{u,t}$; it cannot recover a relevant item that the generator failed to include. Together, \ourname{} establishes a unified framework for transferring LLM knowledge into generative recommendation at three complementary levels.
\MODa{} introduces global semantic priors through input initialization, \MODb{} enhances internal representation learning through hierarchical semantic injection, and \MODc{} refines final recommendation decisions through MoE-based multi-signal calibration.
These components enable semantic knowledge to contribute throughout the recommendation pipeline, from input initialization and candidate generation to final ranking.

\begin{figure}[t]
    \centering
    \includegraphics[width=\linewidth]{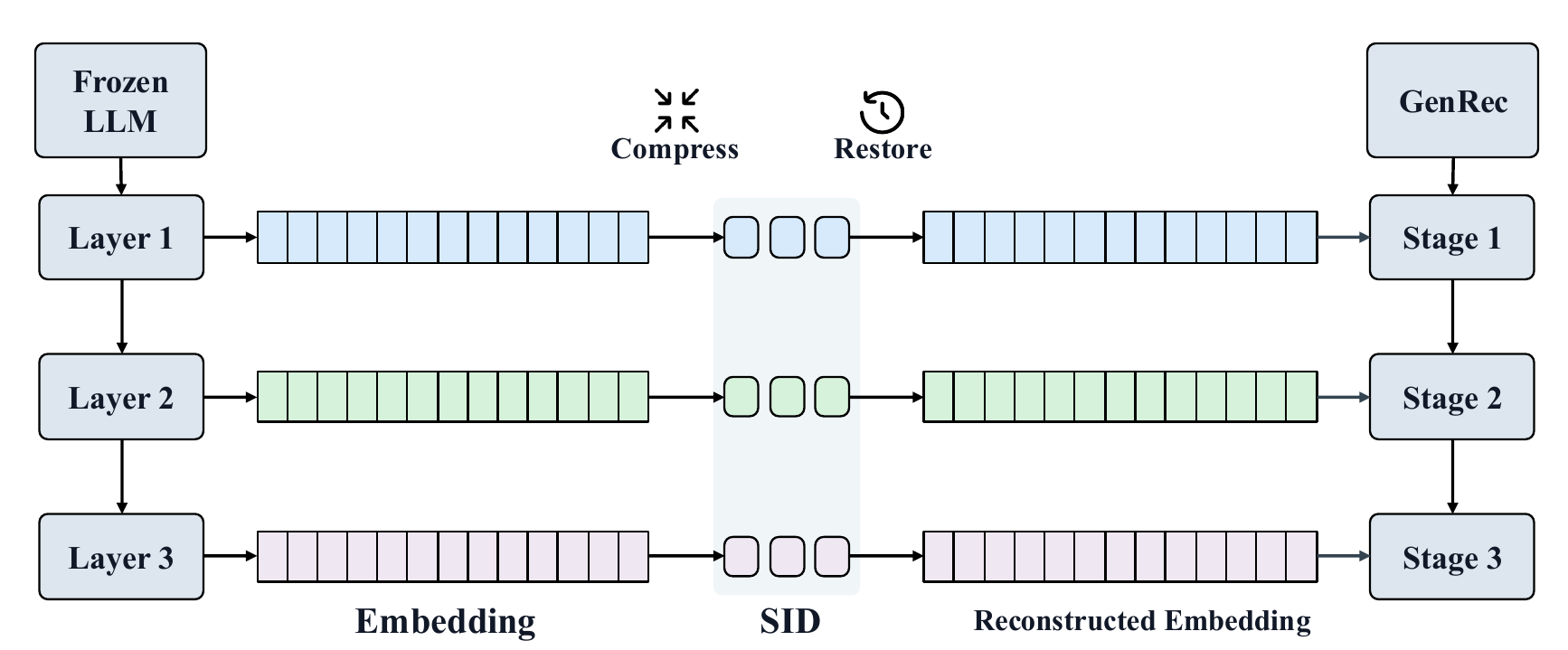}
    \caption{
SID-based compression for hierarchical transfer. Multi-level LLM embeddings are compressed into SIDs and reconstructed for stage-wise semantic transfer.
}
\vskip -0.1in
    \label{fig:HiGenTrans_Industry}
\end{figure}

\section{Industry Deployment}

Deploying \ourname{} at industrial scale introduces additional storage and bandwidth costs, as HRT utilizes semantic embeddings from multiple LLM layers rather than only the final layer. To address this challenge, we adopt the SID-based compression scheme illustrated in Fig.~\ref{fig:HiGenTrans_Industry}. Specifically, hierarchical semantic embeddings are quantized into compact discrete codes and reconstructed through layer-specific quantizers and shared codebooks:

\begin{equation}
\mathbf{v}_{u,t}^{(l)} = Q_l(\mathbf{e}_{u,t}^{(l)}),
\quad
\widehat{\mathbf{e}}_{u,t}^{(l)} = D_l(\mathbf{v}_{u,t}^{(l)}),
\label{eq:sid_compression}
\end{equation}

where $Q_l$ and $D_l$ denote the quantizer and decoder at layer $l$, respectively. Since codebooks are shared across user interaction prefixes, each cached instance stores only a few discrete indices, significantly reducing storage and transmission overhead.

LLM feature extraction and SID assignment are performed offline, while reconstructed embeddings are used during training to mitigate quantization errors. During online inference, lightweight codebook lookups recover semantic features for HRT without LLM execution, enabling efficient and scalable industrial deployment.
\section{Experiments}

We evaluate \ourname{} by addressing four research questions:
\begin{itemize}[leftmargin=*]
    \item \textbf{RQ1:} How does \ourname{} compare with representative sequential and generative recommenders?
    \item \textbf{RQ2:} How much does each transfer module contribute? 
    \item \textbf{RQ3:} Are the gains robust to reduced training data and consistent across user and item segments?
    \item \textbf{RQ4:} Does hierarchical transfer remain effective in industrial recommendation scenarios?
\end{itemize}

\subsection{Experimental Settings}
  \begin{table}[h]
  \centering
  \small
  \setlength{\tabcolsep}{4.0pt}
  \caption{Statistics of the datasets used in our experiments.}
  \vskip -0.1in
  \label{tab:dataset_statistics}
  \begin{tabular}{cccccc}
  \toprule
  \textbf{Dataset}
  & \textbf{\#Users}
  & \textbf{\#Items}
  & \textbf{\#Interactions}
  & \textbf{Density (\%)}
  & \textbf{Avg. Len.} \\
  \midrule
  Beauty & 22,363 & 12,101 & 198,502 & 0.073 & 8.88 \\
  Sports & 35,598 & 18,357 & 296,337 & 0.045 & 8.32 \\
  Toys   & 19,412 & 11,924 & 167,597 & 0.072 & 8.63 \\
  \bottomrule
  \end{tabular}
  \vskip -0.1in
  \end{table}

\begin{table*}[t]
\centering
\small
\setlength{\tabcolsep}{3.2pt}
\renewcommand{\arraystretch}{1.10}
\caption{
Overall performance comparison on three Amazon datasets.
Best results are highlighted in \textbf{bold}, and second-best results are \underline{underlined}.
}
\vskip -0.1in
\label{tab:main_results}
\begin{tabular*}{\textwidth}{
@{\extracolsep{\fill}}
ll
ccccc
ccccc
ccccc
@{}
}
\toprule

\multicolumn{2}{c}{}
& \multicolumn{5}{c}{\textbf{Beauty}}
& \multicolumn{5}{c}{\textbf{Sports}}
& \multicolumn{5}{c}{\textbf{Toys}} \\

\cmidrule(lr){3-7}
\cmidrule(lr){8-12}
\cmidrule(lr){13-17}

\textbf{Paradigm}
& \textbf{Model}
& \textbf{R@1} & \textbf{R@5} & \textbf{R@10} & \textbf{N@5} & \textbf{N@10}
& \textbf{R@1} & \textbf{R@5} & \textbf{R@10} & \textbf{N@5} & \textbf{N@10}
& \textbf{R@1} & \textbf{R@5} & \textbf{R@10} & \textbf{N@5} & \textbf{N@10} \\

\midrule

\textbf{Recurrent}
& GRU4Rec
& .0125 & .0343 & .0512 & .0235 & .0290
& .0043 & .0160 & .0262 & .0102 & .0134
& .0091 & .0239 & .0358 & .0167 & .0205 \\

\midrule

\multirow{3}{*}{\textbf{Attention-based}}
& SASRec
& .0077 & .0566 & .0880 & .0327 & .0428
& .0046 & .0307 & .0492 & .0180 & .0239
& .0059 & .0620 & \underline{.0927} & .0352 & .0450 \\

& HSTU
& .0118 & .0563 & .0868 & .0346 & .0445
& .0076 & .0260 & .0427 & .0167 & .0220
& .0094 & \underline{.0630} & .0919 & .0370 & .0464 \\

& BERT4Rec
& .0050 & .0230 & .0413 & .0138 & .0197
& .0033 & .0147 & .0259 & .0090 & .0126
& .0043 & .0241 & .0414 & .0139 & .0194 \\

\midrule

\multirow{3}{*}{\textbf{Generative Rec.}}
& TIGER
& .0113 & .0391 & .0626 & .0253 & .0329
& .0077 & .0250 & .0409 & .0164 & .0216
& .0128 & .0360 & .0565 & .0245 & .0311 \\

& RPG
& .0196 & .0506 & .0732 & .0354 & .0427
& .0104 & .0280 & .0420 & .0194 & .0239
& .0199 & .0557 & .0817 & .0382 & .0465 \\

& GRLM
& \underline{.0251} & \underline{.0634} & \underline{.0893}
& \underline{.0447} & \underline{.0530}
& \underline{.0142} & \underline{.0388} & \underline{.0562}
& \underline{.0267} & \underline{.0323}
& \underline{.0241} & .0603 & .0842
& \underline{.0428} & \underline{.0504} \\

\midrule

\rowcolor{lightgray}
\textbf{Ours}
& \textbf{HiGenTrans}
& \textbf{.0292} & \textbf{.0677} & \textbf{.0917}
& \textbf{.0489} & \textbf{.0566}
& \textbf{.0147} & \textbf{.0422} & \textbf{.0594} & \textbf{.0287} & \textbf{.0342}
& \textbf{.0372} & \textbf{.0818} & \textbf{.1047}
& \textbf{.0600} & \textbf{.0674} \\

\bottomrule
\end{tabular*}
\end{table*}

\subsubsection{\textbf{Datasets and Protocol}}
We conduct experiments on three widely used categories from the Amazon review benchmark: \textit{Beauty}, \textit{Sports and Outdoors} (\textit{Sports}), and \textit{Toys and Games} (\textit{Toys}). As summarized in Table~\ref{tab:dataset_statistics}, Beauty contains 22,363 users, 12,101 items, and 198,502 interactions; Sports contains 35,598 users, 18,357 items, and 296,337 interactions; and Toys contains 19,412 users, 11,924 items, and 167,597 interactions. We order each user's interactions chronologically and adopt a leave-one-out protocol: the last interaction is used for testing, the penultimate interaction for validation, and all preceding interactions for training. We construct next-item prediction instances from the training prefixes and retain at most the 50 most recent interactions.
Both the LLM input and the GenRec input are truncated at the same prediction boundary; neither contains the validation/test target or any later interaction.
This protocol ensures that every semantic representation is computed from information observable before the target interaction.

\subsubsection{\textbf{Baselines}}
We compare \ourname{} with representative methods from three recommendation paradigms. \textbf{GRU4Rec}~\cite{hidasi2015session} is a recurrent sequential recommender. \textbf{SASRec}~\cite{kang2018self}, \textbf{HSTU}~\cite{zhai2024actions}, and \textbf{BERT4Rec}~\cite{sun2019bert4rec} are attention-based models that learn sequential dependencies through self-attention, hierarchical transduction, and bidirectional masked-item prediction, respectively.
The generative baselines include \textbf{TIGER}~\cite{rajput2023recommender}, which autoregressively generates hierarchical SIDs, \textbf{RPG}~\cite{hou2025generating}, which generates long SIDs in parallel, and \textbf{GRLM}~\cite{zhang2026unleashing}, our generative backbone.

\subsubsection{\textbf{Implementation and Evaluation}}
We build \ourname{} on the GRLM generative backbone and use Qwen3-Embedding-8B as the frozen semantic encoder. Each item is represented by a three-level SID obtained through residual K-means quantization. Item embeddings are projected to 256 dimensions and quantized with three codebooks of size 8,192, yielding the discrete tokens used for autoregressive generation. All semantic features are extracted only from the observable history prefix and can be cached offline.
We optimize the trainable modules with AdamW ($\beta_1=0.9$, $\beta_2=0.95$) in \texttt{bfloat16}, using a batch size of 32, a maximum sequence length of 1,280, and gradient clipping at 1.0. The learning rate follows a cosine schedule from $1\times10^{-4}$ to $1\times10^{-6}$ after 50 warm-up steps. Hyperparameters are selected on the validation set. We report Recall@1/5/10 (R@1/5/10) and NDCG@5/10 (N@5/10). For generative methods, decoded SIDs are mapped to item IDs; when multiple items share an SID, all corresponding items are included in evaluation.

\subsection{Overall Performance (RQ1)}

Table~\ref{tab:main_results} reports performance on the three public datasets. \ourname{} achieves the best result for every dataset and metric. On Beauty, it obtains 0.0917 R@10 and 0.0566 N@10, improving over the strongest baselines by 2.7\% and 6.8\%, respectively. On Sports, it reaches 0.0594 R@10 and 0.0342 N@10, corresponding to gains of 5.7\% and 5.9\%. The largest improvements occur on Toys, where \ourname{} achieves 0.1047 R@10 and 0.0674 N@10, exceeding the best competing results by 12.9\% and 33.7\%.

The gains are particularly strong at the top of the ranking. On Beauty, \ourname{} improves GRLM from 0.0251 to 0.0292 in R@1 (16.3\%) and from 0.0447 to 0.0489 in N@5 (9.4\%). On Sports, R@5 rises from 0.0388 to 0.0422 (8.8\%), while N@5 increases from 0.0267 to 0.0287 (7.5\%). On Toys, the strongest competing R@1 is 0.0241, whereas \ourname{} reaches 0.0372, a 54.4\% relative improvement. It also raises the best R@5 from 0.0630 to 0.0818 (29.8\%) and the best N@5 from 0.0428 to 0.0600 (40.2\%). These improvements across recall and position-sensitive NDCG indicate that the proposed transfer modules help both retrieve relevant items and place them nearer the top of the recommendation list.

The comparison also reveals that no single baseline paradigm dominates all settings. GRLM is the strongest baseline on Beauty and Sports, whereas attention-based SASRec and HSTU remain competitive on Toys. \ourname{} improves over both groups, suggesting that LLM-derived semantics complement rather than replace the collaborative patterns learned from interaction sequences. The relative N@10 gain is larger than the R@10 gain on every dataset (6.8\% versus 2.7\% on Beauty, 5.9\% versus 5.7\% on Sports, and 33.7\% versus 12.9\% on Toys). Since NDCG is more sensitive to rank positions, this gap indicates that the framework improves ordering in addition to retrieval. The component analysis below attributes part of this effect directly to MDC.

\begin{table}[t]
\centering
\caption{Ablation study of HiGenTrans on the Toys dataset.}
\label{tab:ablation}
\vskip -0.1in
\begin{tabular}{lcccc}
\toprule
Method & R@1 & R@5 & N@10 & R@10 \\
\midrule
Base LLM4GenRec     & .0229 & .0581 & .0480 & .0803 \\
+Global Semantic Priming & .0269 & .0696 & .0569 & .0958 \\
+Hierarchical Transfer  & .0269 & .0707 & .0577 & .0967 \\
+Decision Calibration    & .0319 & .0753 & .0628 & .1010 \\
\midrule
\rowcolor{lightgray}
Full Model & \textbf{.0372} & \textbf{.0818} & \textbf{.0674} & \textbf{.1047} \\

\bottomrule
\end{tabular}
\vskip -0.1in
\end{table}

\subsection{Component Analysis (RQ2)}

Table~\ref{tab:ablation} isolates the three transfer levels on Toys by adding each module separately to the same LLM4GenRec base. Every module improves all reported metrics. GSP increases R@10 from 0.0803 to 0.0958 and N@10 from 0.0480 to 0.0569, absolute gains of 1.55 and 0.89 percentage points, respectively. Thus, a single prefix-level semantic prior already compensates for information absent from discrete interaction IDs. HRT reaches 0.0967 R@10 and 0.0577 N@10, slightly outperforming GSP at the deeper cutoffs. Because the two variants use the same semantic source and differ mainly in transfer placement, the result supports preserving depth-specific features instead of relying only on input propagation. MDC is the strongest individual component, reaching 0.1010 R@10 and 0.0628 N@10. Its larger NDCG gain is consistent with its objective: it cannot introduce a missing beam candidate, but it can promote an already generated target into the evaluated prefix of the list.

Combining all three modules yields the best results: 0.0372 R@1, 0.0818 R@5, 0.1047 R@10, and 0.0674 N@10. Relative to the base model, the full framework improves these metrics by 62.4\%, 40.8\%, 30.4\%, and 40.4\%, respectively. It also outperforms MDC, the strongest single-module variant, by 16.6\% in R@1 and 7.3\% in N@10. This improvement over every single-module model is important: the generator-side modules and the reranker are not substitutes. GSP and HRT change the representations used to form the beam, while MDC uses evidence unavailable to raw beam scores to determine which of those candidates occupy the top positions.

\subsection{Stage-wise Discriminability of LLM Representations (RQ2)}

\begin{table}[t]
    \centering
    \caption{
    Stage-wise discriminative ability of LLM representations
    measured by candidate-ranking AUC.
    $\Delta$ indicates absolute AUC changes in percentage points.
    }
    \label{tab:stage_discriminability}
    \vskip -0.1in
    \setlength{\tabcolsep}{3.5pt}
    \renewcommand{\arraystretch}{1.12}
    \resizebox{\columnwidth}{!}{
    \begin{tabular}{lcccccccc}
        \toprule
        & \multicolumn{4}{c}{\textbf{Candidate-ranking AUC}}
        & \multicolumn{3}{c}{\textbf{Stage-wise Gain}}
        & \textbf{Overall} \\
        \cmidrule(lr){2-5}
        \cmidrule(lr){6-8}
        \cmidrule(lr){9-9}
        \textbf{Dataset}
        & \textbf{S1} & \textbf{S2}
        & \textbf{S3} & \textbf{S4}
        & $\Delta_{2-1}$
        & $\Delta_{3-2}$
        & $\Delta_{4-3}$
        & $\Delta_{4-1}$ \\
        \midrule
        Toys
        & 0.5788 & 0.5863 & \textbf{0.5956} & 0.5953
        & +0.75 & +0.93 & -0.03 & +1.65 \\
        Beauty
        & 0.5434 & 0.5414 & 0.5475 & \textbf{0.5530}
        & -0.20 & +0.61 & +0.55 & +0.96 \\
        Sports
        & 0.5302 & 0.5335 & \textbf{0.5481} & 0.5440
        & +0.33 & +1.46 & -0.41 & +1.38 \\
        \bottomrule
    \end{tabular}
    }
    \vskip -0.1in
\end{table}

To investigate how semantic discriminability evolves
across different stages of the LLM, we evaluate the
candidate-ranking capability of representations extracted
from four stages (Stage 1--4).
Specifically, we use cosine similarity between candidate
items and user historical representations as the ranking
score, and measure their ability to distinguish ground-truth
items from negative candidates using AUC.
The evaluation considers users whose ground-truth items
appear among the 20 generated candidates and have
available LLM representations.
As shown in Table~\ref{tab:stage_discriminability},
deeper stages generally exhibit stronger discriminative
ability across all three datasets.
Compared with Stage 1, Stage 3 improves AUC by
0.0168, 0.0041, and 0.0179 on Toys, Beauty, and Sports,
respectively, suggesting that deeper LLM representations
capture more recommendation-relevant semantic information.
However, Stage 4 does not consistently outperform Stage 3,
indicating that semantic discriminability tends to
saturate at deeper stages.
These findings suggest that deeper stages provide stronger
standalone ranking signals, motivating hierarchical
semantic transfer across multiple representation stages
without requiring dataset-specific stage selection.

\subsection{Data-Efficiency Analysis (RQ3)}

\begin{table}[t]
    \centering
    \caption{
    Robustness to reduced training data on Beauty.
    }
    \label{tab:training_user_ratio}
    \vskip -0.1in
    \setlength{\tabcolsep}{6pt}
    \renewcommand{\arraystretch}{1.12}
    \begin{tabular}{lcccc}
        \toprule
        \textbf{Users}
        & \textbf{Hit@10}
        & \textbf{Gain (\%)}
        & \textbf{Range (\%)}
        & $\boldsymbol{\Delta_{100}}$ \textbf{(pp)} \\
        \midrule
        \rowcolor{lightgray}
        100\%
        & \textbf{0.0917}
        & \textbf{+8.86}
        & ---
        & --- \\
        \midrule
        95\%
        & 0.0908
        & +7.84
        & 7.17--8.44
        & $-$0.09 $\downarrow$\\
        90\%
        & 0.0906
        & +7.59
        & 7.48--7.70
        & $-$0.11 $\downarrow$\\
        80\%
        & 0.0903
        & +7.24
        & 6.58--8.01
        & $-$0.14 $\downarrow$ \\
        \bottomrule
    \end{tabular}
    \vskip -0.1in
\end{table}

Table~\ref{tab:training_user_ratio} evaluates \ourname{} after subsampling the training users on Beauty. Reducing the training set from 100\% to 95\%, 90\%, and 80\% changes R@10 from 0.0917 to 0.0908, 0.0906, and 0.0903, respectively. Even at the 80\% setting, the absolute decrease is only 0.0014 (1.5\% relative to the full-data result), while the improvement over the correspondingly trained base model remains 7.24\%. The relative gain varies by only 1.62 percentage points across the four ratios. This stability is consistent with the role of a frozen semantic encoder: when interaction supervision decreases, the external semantic features remain available.

\subsection{Segment-Level Benefits (RQ3)}
\label{sec:segment_analysis}

\begin{table}[t]
\centering
\small
\setlength{\tabcolsep}{4pt}
\renewcommand{\arraystretch}{1.10}
\caption{
Performance across item popularity groups on Toys.
Gains denote relative improvements (\%) over the baseline.
Darker green indicates larger gains.
}
\vskip -0.1in
\label{tab:segment_toys_popularity}
\resizebox{\columnwidth}{!}{%
\begin{tabular}{lcccc}
\toprule
\textbf{Popularity}
& \multicolumn{2}{c}{\textbf{Hit@10}}
& \multicolumn{2}{c}{\textbf{NDCG@10}} \\
\cmidrule(lr){2-3}
\cmidrule(lr){4-5}
& \textbf{Base / Ours}
& \textbf{Gain}
& \textbf{Base / Ours}
& \textbf{Gain} \\
\midrule
\rowcolor{lightgray}
All
& .0932 / .1047
& {+12.3\%}
& .0544 / .0674
& {+23.9\%} \\
\midrule
$\leq$ 2
& .0330 / .0400
& \gaincell{4}{+21.3\%}
& .0239 / .0341
& \gaincell{4}{+42.8\%} \\
3--5
& .0591 / .0770
& \gaincell{5}{\textbf{+30.4\%}}
& .0327 / .0536
& \gaincell{5}{\textbf{+64.1\%}} \\
6--15
& .0964 / .1097
& \gaincell{3}{+13.7\%}
& .0575 / .0690
& \gaincell{2}{+20.2\%} \\
$\geq$ 16
& .1341 / .1412
& \gaincell{1}{+5.2\%}
& .0768 / .0867
& \gaincell{1}{+13.0\%} \\
\bottomrule
\end{tabular}%
}
\vskip -0.1in
\end{table}

\begin{table}[t]
\centering
\small
\setlength{\tabcolsep}{4pt}
\renewcommand{\arraystretch}{1.10}
\caption{
Performance across user history lengths on Toys.
Gains denote relative improvements (\%) over the baseline.
Darker green indicates larger gains.
}
\vskip -0.1in
\label{tab:segment_toys_history}
\resizebox{\columnwidth}{!}{%
\begin{tabular}{lcccc}
\toprule
\textbf{History}
& \multicolumn{2}{c}{\textbf{Hit@10}}
& \multicolumn{2}{c}{\textbf{NDCG@10}} \\
\cmidrule(lr){2-3}
\cmidrule(lr){4-5}
& \textbf{Base / Ours}
& \textbf{Gain}
& \textbf{Base / Ours}
& \textbf{Gain} \\
\midrule
\rowcolor{lightgray}
All
& .0932 / .1047
& {+12.3\%}
& .0544 / .0674
& {+23.9\%} \\
\midrule
3
& .1023 / .1138
& \gaincell{2}{+11.3\%}
& .0612 / .0736
& \gaincell{2}{+20.3\%} \\
4--5
& .1001 / .1108
& \gaincell{2}{+10.7\%}
& .0582 / .0719
& \gaincell{2}{+23.6\%} \\
6--9
& .0739 / .0849
& \gaincell{3}{+14.9\%}
& .0419 / .0554
& \gaincell{3}{\textbf{+32.4\%}} \\
$\geq$ 10
& .0841 / .0978
& \gaincell{4}{\textbf{+16.3\%}}
& .0479 / .0598
& \gaincell{3}{+25.0\%} \\
\bottomrule
\end{tabular}%
}
\vskip -0.1in
\end{table}
\noindent\textbf{Impact of Item Popularity.}
We evaluate \ourname{} across different item-popularity groups on Toys, as shown in Table~\ref{tab:segment_toys_popularity}. Our method consistently outperforms the baseline across all groups, demonstrating its effectiveness for both head and long-tail items. Notably, the largest improvements occur for items with popularity 3--5, achieving relative gains of 30.4\% in Hit@10 and 64.1\% in NDCG@10. Even for extremely long-tail items (popularity $\leq 2$), \ourname{} achieves gains of 21.3\% and 42.8\%, respectively. These results highlight the effectiveness of semantic knowledge transfer in alleviating item sparsity.

\noindent\textbf{Impact of User History Length.}
We further investigate performance across different user history lengths in Table~\ref{tab:segment_toys_history}. \ourname{} consistently improves recommendation accuracy across all groups, with relative Hit@10 gains ranging from 10.7\% to 16.3\%. Users with longer histories ($\geq 10$) achieve the largest Hit@10 improvement of 16.3\%, while users with 6--9 interactions obtain the highest NDCG@10 gain of 32.4\%. These findings demonstrate that our method effectively leverages semantic knowledge across varying levels of user interaction sparsity.

\subsection{Candidate Ranking Correction (RQ3)}

\begin{figure}
    \centering
    \includegraphics[width=\linewidth]{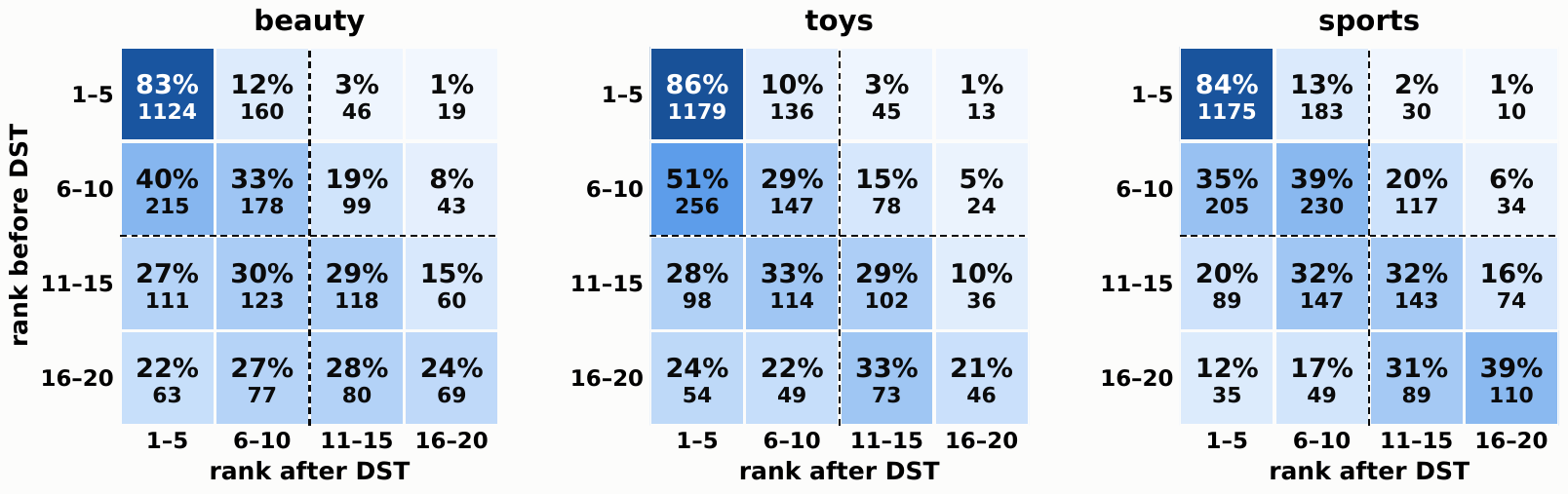}
    \caption{
    Ground-truth ranking transitions before and after MDC.
    Each cell shows the transition percentage and user count.
    Dashed lines indicate the top-10 boundary.
    }
    \vskip -0.1in
    \label{fig:rank_correction}
\end{figure}
To evaluate the effectiveness of MDC, we analyze the ranking transitions of ground-truth items within the top-20 candidates, as shown in Fig.~\ref{fig:rank_correction}. MDC successfully promotes 53.4\%, 55.1\%, and 43.5\% of items originally ranked 11--20 into the top-10 on Beauty, Toys, and Sports, respectively. Meanwhile, 83--86\% of top-5 items retain their positions within the top-5. These results demonstrate that MDC effectively corrects suboptimal rankings while preserving high-quality predictions.

\subsection{Industrial Evaluation (RQ4)}

\begin{table}[t]
\centering
\setlength{\tabcolsep}{3.5pt}
\renewcommand{\arraystretch}{1.10}
\caption{
SID prediction performance across different hierarchical
levels in two industrial cross-domain recommendation scenarios.
$\Delta$ denotes absolute improvements, with darker green
indicating larger gains.
}
\label{tab:sid_performance}
\vskip -0.1in
\begin{tabular}{@{}lccc@{\hspace{5pt}}ccc@{}}
\toprule
& \multicolumn{3}{c}{\textbf{Domain 1}}
& \multicolumn{3}{c}{\textbf{Domain 2}} \\
\cmidrule(lr){2-4}
\cmidrule(lr){5-7}

\textbf{Method}
& \textbf{SID$_1$}
& \textbf{SID$_2$}
& \textbf{SID$_3$}
& \textbf{SID$_1$}
& \textbf{SID$_2$}
& \textbf{SID$_3$} \\
\midrule

Baseline
& 0.1360
& 0.4789
& 0.8038
& 0.2149
& 0.4698
& 0.6083 \\

\rowcolor{lightgray}
Ours
& \textbf{0.6178}
& \textbf{0.6155}
& \textbf{0.8226}
& \textbf{0.8347}
& \textbf{0.5797}
& \textbf{0.6383} \\

\midrule

$\Delta$
& \gaincell{5}{+0.4818}
& \gaincell{3}{+0.1366}
& \gaincell{1}{+0.0188}
& \gaincell{5}{+0.6198}
& \gaincell{3}{+0.1099}
& \gaincell{1}{+0.0300} \\

\bottomrule
\end{tabular}
\vskip -0.1in
\end{table}

We further evaluate hierarchical transfer in two real-world cross-domain e-commerce scenarios. As shown in Table~\ref{tab:sid_performance}, \ourname{} consistently improves SID prediction accuracy across all hierarchical levels in both domains. Notably, SID$_1$ achieves substantial absolute gains of 48.18 and 61.98 percentage points, while SID$_2$ and SID$_3$ show smaller but consistent improvements. This trend suggests that LLM semantic knowledge is particularly beneficial for coarse-grained predictions, while fine-grained predictions rely more on collaborative signals. Overall, these results validate the effectiveness of hierarchical transfer in industrial recommendation systems.
\section{Conclusion}
In this work, we proposed \ourname{}, a hierarchical LLM-to-GenRec knowledge transfer framework that enhances generative recommendation at the input, representation, and decision levels. Specifically, GSP introduces global semantic priors, HRT transfers multi-level LLM representations into the generative backbone, and MDC integrates complementary semantic, generative, and contextual signals to refine candidate rankings. To support efficient deployment, we further incorporate offline semantic feature compression, avoiding online LLM inference. Extensive experiments on three public benchmarks demonstrate consistent improvements, achieving up to 12.94\% relative gains in Recall@10. 
Overall, our findings highlight the potential of hierarchical semantic knowledge transfer and decision-level calibration for building accurate and efficient generative recommender systems.

\bibliographystyle{ACM-Reference-Format}
\balance
\bibliography{citation}

\end{document}